\documentclass[runningheads]{llncs}
\usepackage{graphicx}

\usepackage{tikz}
\usepackage{comment}
\usepackage{amsmath,amssymb} 
\usepackage{color}

\usepackage[pagebackref,breaklinks,colorlinks]{hyperref}
\usepackage{multirow}
\usepackage[normalem]{ulem}
\useunder{\uline}{\ul}{}

\usepackage{pbox}

\usepackage{amssymb}
\usepackage{wrapfig}
\usepackage{algorithm}
\usepackage[noend]{algcompatible}

\usepackage[accsupp]{axessibility}  

\usepackage{wrapfig}

\newcommand{\nickname}{RangeUDF}

\begin{document}
\pagestyle{headings}
\mainmatter
\def\ECCVSubNumber{7056}  

\title{\nickname{}: Semantic Surface Reconstruction from 3D Point Clouds} 

\titlerunning{RangeUDF}
%
\author{Bing Wang$^{1,2}$, Zhengdi Yu$^3$, Bo Yang$^{2\thanks{Corresponding author}}$, Jie Qin$^4$, Toby Breckon$^3$, \\ Ling Shao$^5$, Niki Trigoni$^1$, Andrew Markham$^1$ \\
\vspace{0.2cm}
\normalfont
\normalsize\textsuperscript{$1$}University of Oxford\quad
\normalsize\textsuperscript{$2$}The Hong Kong Polytechnic University\quad 
\normalsize\textsuperscript{$3$}Durham University\quad
\normalsize\textsuperscript{$4$}Nanjing University of Aeronautics and Astronautics\quad 
\normalsize\textsuperscript{$5$}IIAI\quad}
\authorrunning{B. Wang et al.}
%
\institute{}

\maketitle
\vspace{-0.2cm}
\begin{abstract}
We present \nickname{}, a new implicit representation based framework to recover the geometry and semantics of continuous 3D scene surfaces from point clouds. Unlike occupancy fields or signed distance fields which can only model closed 3D surfaces, our approach is not restricted to any type of topology. Being different from the existing unsigned distance fields, our framework does not suffer from any surface ambiguity. In addition, our \nickname{} can jointly estimate precise semantics for continuous surfaces. The key to our approach is a range-aware unsigned distance function together with a surface-oriented semantic segmentation module. Extensive experiments show that \nickname{} clearly surpasses state-of-the-art approaches for surface reconstruction on four point cloud datasets. Moreover, \nickname{} demonstrates superior generalization capability across multiple unseen datasets, which is nearly impossible for all existing approaches. The code is available at \url{https://github.com/vLAR-group/RangeUDF}.

\end{abstract}

\section{Introduction}
Recovering fine-grained geometry and the semantic composition of 3D scene point clouds is a key enabler for many cutting-edge applications in augmented reality and robotics. To obtain geometric details, classical methods \cite{Kazhdan2013} usually rely on strong geometric priors such as local linearity, resulting in the recovered surfaces to be over-smooth, losing fine details. 

By encoding geometry into multi-layer perceptrons, recent implicit representations have shown great potential to reconstruct complex shapes from point clouds and images \cite{Park2019,Peng2020a,Mildenhall2020}. Their key advantage is the ability to represent 3D structures as continuous functions, which can achieve unlimited spatial resolution in theory. Implicit representations can broadly be divided into: 1) occupancy fields (OF) \cite{Mescheder2019}, 2) signed distance fields (SDF) \cite{Park2019}, 3) radiance fields (NeRF) \cite{Mildenhall2020}, and 4) hybrid representations \cite{Oechsle2021}. Although they have been successfully applied and achieved impressive results in 1) image based shape reconstruction \cite{Chibane2020,Sucar2021}, 2) image based scene understanding \cite{Zhang2021b}, 3) differentiable rendering \cite{Niemeyer2019,Liu2019b}, 4) novel view synthesis \cite{Trevithick2021}, and 5) shape generation \cite{Niemeyer2021}, few works are able to recover the precise 3D surfaces and semantics of large-scale point clouds such as spacious rooms with dozens of chairs and tables. Fundamentally, this is because the true surfaces of these sparse point clouds are inherently open and have arbitrary topology. However, the widely used OF \cite{Mescheder2019} and SDF \cite{Park2019} can only model closed surfaces. Although NeRF \cite{Mildenhall2020} methods can take point clouds as input to estimate continuous structures, the underlying volume rendering does not provide sufficient geometric constraints to recover fine-grained details. 

\begin{figure}[t]
\centering
   \includegraphics[width=1\linewidth]{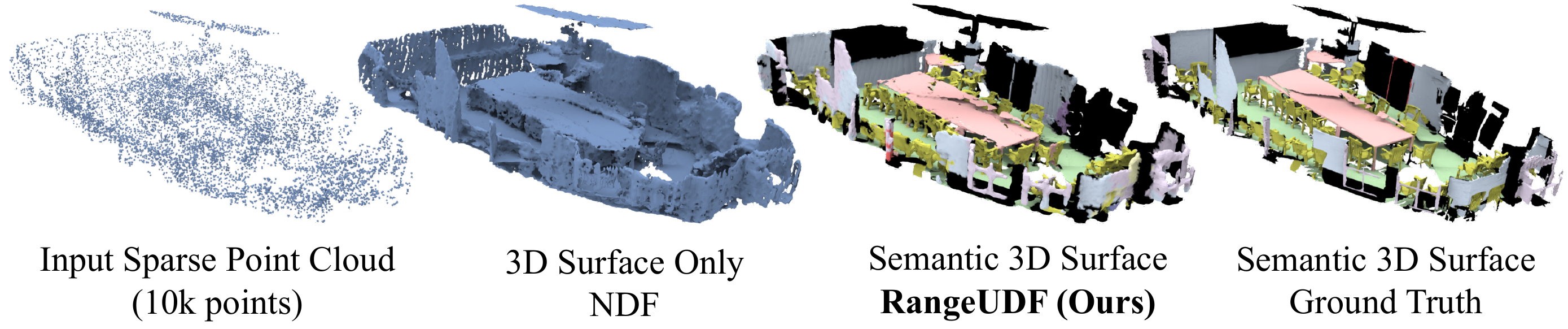}
 \vspace{-0.4cm}
\caption{Given a sparse input point cloud with complex structures from ScanNet \cite{Dai2017}, our \nickname{} jointly recovers precise geometry and accurate semantics of continuous 3D surfaces, while existing methods such as NDF \cite{Chibane2020a} cannot.}
\label{fig:opening}
\vspace{-0.2cm}
\end{figure}

This lacuna in modeling open surfaces has recently begun to be filled by a handful of works. Among them, there are two pipelines. The first pipeline is SAL \cite{Atzmon2020} and its variant \cite{Atzmon2020a}. By learning with an unsigned objective and careful initialization of an implicit decoder, they do not require closed shapes in training. However, their final recovered 3D surfaces tend to be closed, leading to missing interior structures. The second pipeline is based on NDF method \cite{Chibane2020a}. Given an input point cloud and an arbitrary query point in 3D space, they learn to directly regress the unsigned distance between that query point and the underlying surface. Albeit simple, NDF achieves high-fidelity results in recovering truly open surfaces, especially for object-level and small-scale dense point clouds. 

Nevertheless, NDF cannot be efficiently extended to scene-level point clouds due to two limitations. 1) Since NDF is based on voxel representations to extract local features, it requires high memory usage or time-consuming sliding windows to process large point clouds, and often fails to retain fine-grained details given limited voxel resolutions. 2) To infer the unsigned distance for a specific query point, NDF adopts trilinear interpolation to compute a feature vector for that query point from its neighbouring voxels. However, this key step is likely to suffer from surface ambiguity when the input point clouds are sparse and with variable density. These two limitations directly lead to the estimated unsigned distances inaccurate, and the recovered surfaces over-smooth. In addition, NDF does not simultaneously estimate surface semantics, and it is unclear how to integrate valid semantic segmentation for query points that might be on or off surfaces.


In this paper, we introduce \textbf{range}-aware \textbf{u}nsigned \textbf{d}istance \textbf{f}ields, named \textbf{RangeUDF}, a simple end-to-end neural implicit function that can jointly estimate precise 3D surface structures and semantics from raw and large-scale point clouds, without suffering from the limitations of existing approaches. In particular, our framework consists of three major components: 1) a per-point feature extractor that can take large-scale point clouds in a single forward pass, 2) a range-aware neural interpolation module that can clearly avoid the surface ambiguity for all query points to infer accurate unsigned surface distances, and 3) a surface-oriented semantic segmentation module that can effectively learn surface semantics even if the query points are far from the surfaces during training. 

Being built on the above components, in particular the latter two, our method is not restricted to any surface topology, and can recover fine-grained geometry and semantics regardless of the openness, sparsity and density of input point clouds. Our method clearly surpasses the state-of-the-art surface reconstruction approaches on four datasets. In addition, our \nickname{} demonstrates remarkable generalization capability across multiple unseen datasets. Figure~\ref{fig:opening} shows qualitative results of our approach in ScanNet \cite{Dai2017}. Our key contributions are:
\begin{itemize}
    \item We propose a range-aware feature interpolation module to obtain a unique feature vector for each query 3D point. This allows to infer precise unsigned distances without any surface ambiguity.
    \item We introduce a surface-oriented semantic segmentation module that enables our framework to jointly estimate surface semantics.
    \item We demonstrate significant improvement over baselines and surpass the state-of-the-art methods by large margins on four point cloud datasets.
\end{itemize}

Nevertheless, it is worthwhile highlighting that our \nickname{} is technically very simple, intuitive and easy to implement as detailed in Section \ref{sec:method}, while achieving extraordinary performance. Therefore, the core novelty of our method lies in the simplicity and usefulness. We urge the reader to appreciate the neatness instead of expecting complicated or difficult novelties. 

\section{Related Work}

Shape reconstruction has been studied for decades. Classical approaches to recover 3D structures from images mainly include SfM \cite{Ozyesil2017} and SLAM \cite{Cadena2016} systems such as Colmap \cite{Schonberger2016} and ORB-SLAM \cite{Mur-Artal2015}. Surface reconstruction of 3D point clouds mainly relies on global or local smoothness priors such as Poisson reconstruction \cite{Kazhdan2013}, radial basis functions \cite{Carr2001} and moving least-squares surfaces \cite{Guennebaud2007}. A comprehensive survey of classical methods can be found in \cite{Berger2017a}. Recent learning based approaches for 3D shape representation and semantic segmentation are outlined below.

\textbf{Explicit 3D Representations:}
To model explicit 3D geometry of objects and scenes, impressive progress has come from recent advances in recovering voxel grids \cite{Chan2016}, octree \cite{Tatarchenko2017}, point clouds \cite{Fan2017}, triangle meshes \cite{Kato2017} and shape primitives \cite{Zou2017} from either images or point clouds. Although they have shown great performance in shape reconstruction \cite{Yang2018,Tang2019}, completion \cite{Song2017}, shape generation \cite{Lin2017a}, and scene understanding \cite{Tulsiani2017c,Gkioxari2019}, the quality of such discrete shape representations are inherently limited by the spatial resolution and memory footprint. As a consequence, they are hard to scale up to complex 3D scenes. 
\begin{figure}[th]
\centering
   \includegraphics[width=1.0\linewidth]{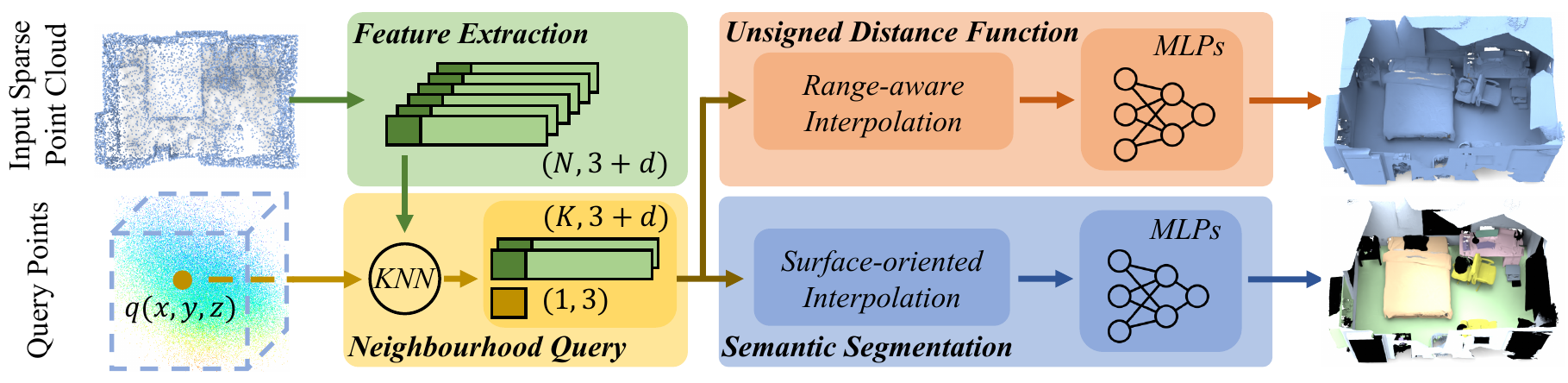}
   \vspace{-0.4cm}
\caption{In our \nickname{}, given an input point cloud, the feature extractor firstly extracts high-quality features for each point. This is followed by our novel range-aware unsigned distance function and surface-oriented segmentation module to learn precise geometry and semantics for each query point.}
\label{fig:architecture}
\vspace{-0.2cm}
\end{figure}

\textbf{Implicit 3D Representations:}
To overcome the discretization issue of explicit representations, MLPs have been recently used to learn implicit functions to represent continuous 3D shapes. Such implicit representations can be generally classified as: 1) occupancy fields \cite{Mescheder2019,Chen2019g}, 2) signed distance fields \cite{Sitzmann2019}, 3) unsigned distance fields \cite{Chibane2020a,Atzmon2020}, 4) radiance fields \cite{Mildenhall2020}, and 5) hybrid fields \cite{Wang2021d}. Among them, both occupancy fields and signed distance fields can only recover closed 3D shapes, while radiance fields focus on neural rendering instead of surface reconstruction. In the past two years, these representations have been extensively studied for shape reconstruction \cite{Peng2020a,Saito2019,Chabra2020,Ma2021,Zhang2021c}, neural rendering and novel view synthesis \cite{Niemeyer2019}, shape generation \cite{Luo2021}, and understanding \cite{Li2020,Zhang2021b}. Although achieving impressive results, almost all of these works focus on single objects or small-scale scenes. In this paper, we scale up the implicit representation to the next level, where our \nickname{} can jointly estimate precise 3D surfaces with semantics from real-world complex point clouds where existing methods cannot.

\textbf{3D Semantic Segmentation:}
To learn per-point semantics for point clouds, existing methods generally include 1) projection and voxel based methods \cite{Graham2018} and 2) point based methods \cite{Qi2016}. Given fully-annotated point cloud datasets, the existing approaches have achieved excellent semantic segmentation accuracy. However, these methods are designed to classify the discrete and individual 3D points explicitly sampled from scene surfaces. With the fast development of implicit representation of 3D scenes, it is desirable to learn semantic information for implicit surfaces. To the best of our knowledge, there is no prior work to jointly estimate structures and semantics for implicit representations from real-world sparse point clouds.

\section{\nickname{}}\label{sec:method}
\subsection{Overview}
Given an input point cloud $\boldsymbol{P}$ of a 3D scene, which consists of $N$ sparsely and non-uniformly distributed 3D points sampled from complex structures and open surfaces, our objective is to reconstruct the underlying continuous surface geometry $\mathcal{S}_{geo}$ and semantic classes $\mathcal{S}_{sem}$. We formulate this problem as learning a neural unsigned distance function $f$ with semantic classification. This neural function takes the entire point cloud $\boldsymbol{P}$ and an arbitrary query point $q$ as input, and then directly predicts the unsigned distance $d_q$ between query point $q$ and the closest surface, together with the semantic label $s_q$ out of $C$ classes for the corresponding closest surface point. Formally, it is defined as below:

\begin{equation}
    (d_q, s_q) =  f(\boldsymbol{P}, q); \quad q\in \mathbb{R}^3, d_q \in \mathbb{R}_0^+, s_q \in \mathbb{R}^C
\end{equation}

As shown in Figure \ref{fig:architecture}, our framework consists of four building blocks: 1) the per-point feature extractor shown in the top-left green block, 2) the query point neighbourhood search module in the bottom-left yellow block, 3) the range-aware unsigned distance function in the top-right orange block, and 4) the surface-oriented semantic segmentation module in the bottom-right blue block. 

For the feature extractor, we simply adopt the existing large-scale-point-cloud friendly RandLA-Net \cite{Hu2020}, although our framework is not restricted to any specific network. For the neighbourhood query module, we use kNN to collect $K$ neighbouring points for every query point $q$ according to point Euclidean distances, although we note that other query methods such as spherical query \cite{Thomas2019} are also applicable.  After collecting $K$ points and their features for each query point $q$, we feed them into our range-aware unsigned distance function and the surface-oriented segmentation module to learn structures and semantics. Details of these two modules are discussed below.

\subsection{Range-aware Unsigned Distance Function}
\begin{figure}[t]
\centering
   \includegraphics[width=0.8\linewidth]{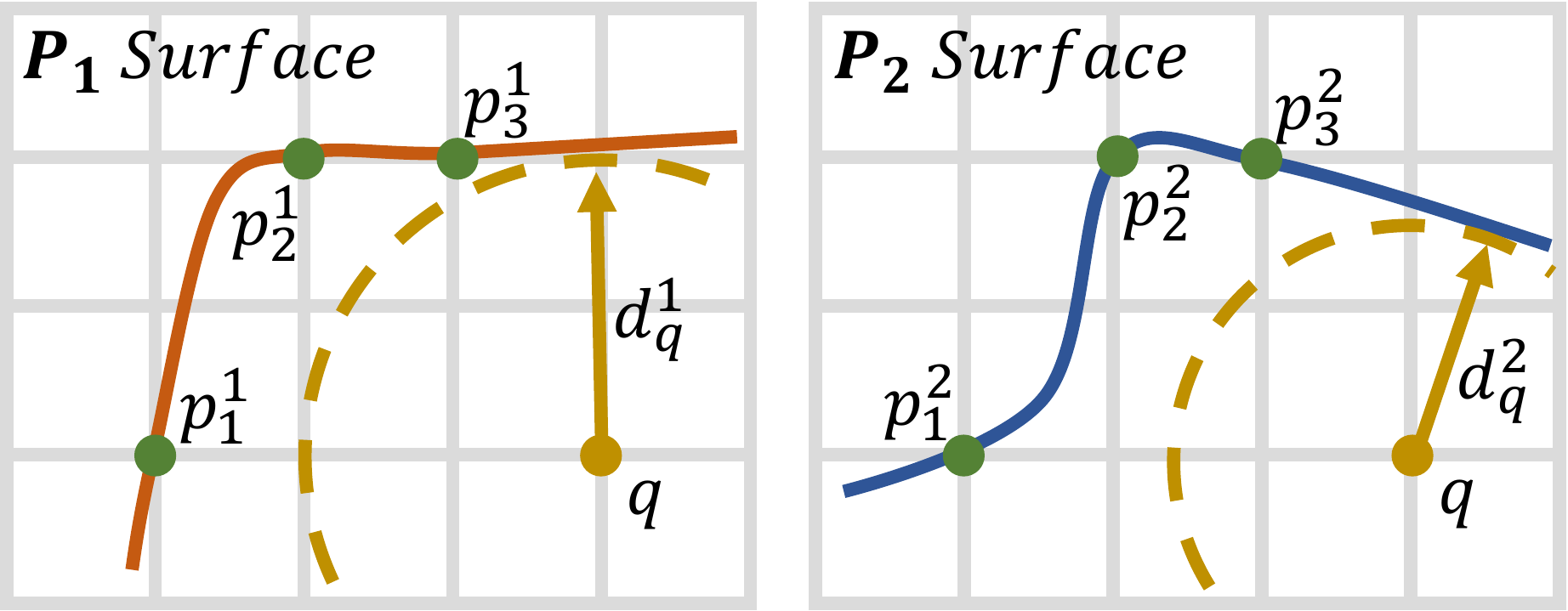}
   \vspace{-0.1cm}
\caption{The ambiguity of simple trilinear interpolation.}
\label{fig:ambiguity_interpolation}
\vspace{-0.2cm}
\end{figure}

\phantom{xxx}\textbf{Ambiguity of Trilinear Interpolation:} 
Given the $K$ neighbouring points and their features for a specific query point $q$, trilinear interpolation is widely used in existing works such as ConvOcc \cite{Peng2020a} and NDF \cite{Chibane2020a} to obtain a weighted feature vector for the query point $q$. However, such simple interpolation may suffer from distance ambiguity during network training when point clouds are sparse with complex structures. As shown in Figure \ref{fig:ambiguity_interpolation}, given two different point clouds $(\boldsymbol{P_1}$, $\boldsymbol{P_2})$ and the same query point $q$ during training, it is very likely that the queried two sets of neighbouring points $\{p_1^1,p_2^1,p_3^1\}$ in $\boldsymbol{P_1}$, and $\{p_1^2, p_2^2, p_3^2\}$ in $\boldsymbol{P_2}$ have the same or similar point locations and features. Naturally, the simple trilinear interpolation will result in a same or similar feature vector for point $q$ in these two scenarios.

However, due to the sparsity and complexity of point clouds $(\boldsymbol{P_1}$, $\boldsymbol{P_2})$, their underlying surfaces, as indicated by the brown and blue lines, can be significantly different. As a result, the ground truth supervision signals, i.e., unsigned distances $d_q^1$ and $d_q^2$, will be quite different. This means that, during training, the network has to predict two vastly different distance values given the same or similar input feature vector of query point $q$. Such ambiguity directly confuses the network during training, and the network tends to predict \textit{mean} distance values. In testing, the network naturally predicts over-smooth surfaces.
\clearpage 

\begin{figure}[h]
\centering
   \includegraphics[width=.85\linewidth]{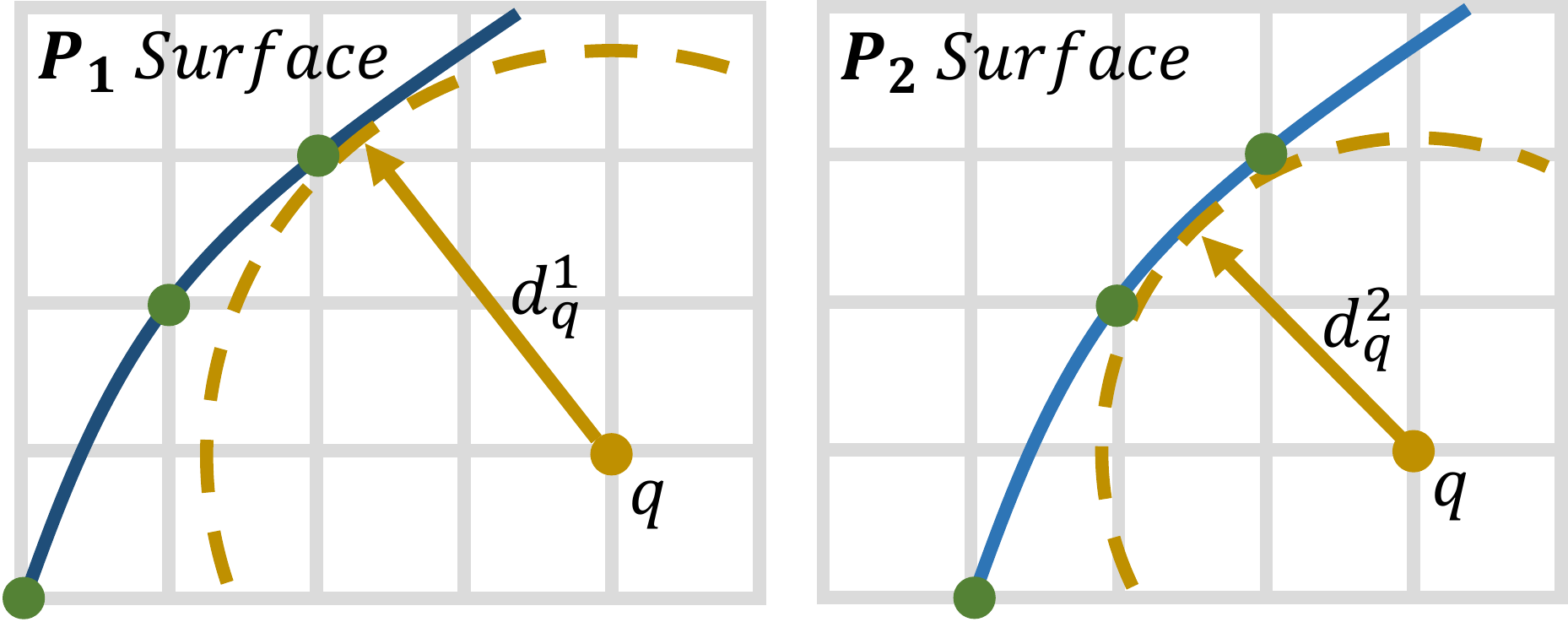}
   \vspace{-0.1cm}
\caption{The importance of relative distance.}
\label{fig:ambiguity_relative}
\vspace{-0.2cm}
\end{figure}

\textbf{Range-aware Neural Interpolation:}
To overcome such ambiguity, we introduce a simple yet effective range-aware neural interpolation module as illustrated in the orange block of Figure \ref{fig:architecture}. In particular, given a query point $q$, we have its neighbouring points $\{p_1 \dots p_K\}$ and point features $\{\boldsymbol{F}_1 \dots \boldsymbol{F}_K\}$ at hand. Our range-aware neural interpolation module explicitly takes into account the relative distances and absolute positions of all neighbouring points. In particular, we encode the range information for each neighbouring point as follows: 
\begin{equation}
    \boldsymbol{R}_k^q = MLP\Big( (q - p_k) \oplus q \oplus p_k \Big)
    \label{eq:range}
\end{equation}

where $q$ and $p_k$ are the $xyz$ positions of points, $\oplus$ is the concatenation operation. For scale consistency, all input point clouds are normalized within a cube of [-0.5, 0.5] along $xyz$ axes in our experiments. As illustrated in Figure \ref{fig:ambiguity_relative}, for the same query point $q$ in space, if the queried neighbouring patches of two point clouds $(\boldsymbol{P_1}$, $\boldsymbol{P_2})$ have similar surfaces but with different position shifts, the relative position term $(q - p_k)$ can directly aid the network to learn the difference between unsigned distances $d_q^1$ and $d_q^2$. Our ablation study in Table \ref{tbl:ablation} clearly shows the effectiveness of such range information.

Unlike trilinear interpolation which simply computes a set of weights $\{w^q_1 \dots  w^q_K\}$ using Euclidean distances between $q$ and $\{p_1 \dots p_k \dots p_K\}$, our module instead learns informative vectors $\{\boldsymbol{R}_1^q \dots \boldsymbol{R}_k^q \dots \boldsymbol{R}_K^q \}$. These are explicitly aware of the range between query point $q$ and all of its neighbouring points, overcoming the distance ambiguity of trilinear interpolation. In order to interpolate a single feature vector $\boldsymbol{F}^q$ for the query point $q$, we concatenate the range vectors with point features followed by a pooling operation. In particular, our neural interpolation is defined as follows:
\begin{equation}
    \boldsymbol{F}^q = \mathcal{A}\Big( [\boldsymbol{R}_1^q \oplus \boldsymbol{F}_1] \dots [\boldsymbol{R}_k^q \oplus \boldsymbol{F}_k] \dots [\boldsymbol{R}_K^q \oplus \boldsymbol{F}_K] \Big)
\end{equation}
where $\mathcal{A}$ is an attention module. We use the simple AttSets \cite{Yang2020} in our experiments, though more advanced modules such as Transformer \cite{Vaswani2017} would likely yield better results.

\textbf{Unsigned Distance Regression:}
In order to infer the final unsigned distance value, we directly feed the feature vector $\boldsymbol{F}^q$ of query point $q$ into a series of MLPs. Identical to NDF \cite{Chibane2020a}, the output layer is followed by a ReLU function, clipping the distance value to be equal/greater than $0$.  

\subsection{Surface-oriented Semantic Segmentation}
\setlength{\columnsep}{10pt}
\begin{wrapfigure}[14]{R}{0.4\textwidth}
\raisebox{5pt}[\dimexpr\height-1.5\baselineskip\relax]{
\centering
\includegraphics[width=1.\linewidth]{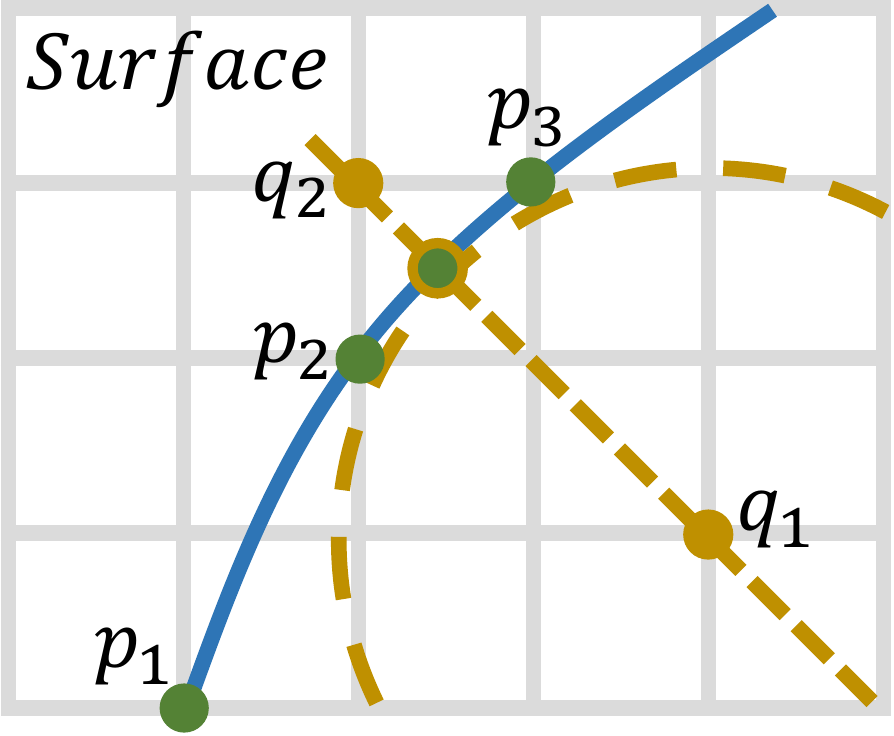}}
\vspace{-0.8cm}
\caption{Eliminating the absolute position of the query point.}
\label{fig:ambiguity_semantic}
\end{wrapfigure}

Being different from unsigned distance estimation, it is non-trivial to learn valid semantic classes for continuous surfaces. The key difference is that, for those query points corresponding to empty space, they do not have any valid semantic labels to supervise. Instead, only the points on surface patches have supervision signals. A na\"ive strategy is to separately optimize unsigned distance branch using both on/off-surface points, while optimizing semantic segmentation branch using on-surface points only. However, such a strategy would inevitably lead to an imbalance and ineffective optimization of two branches as shown in the appendix. 

To overcome this issue, we introduce a surface-oriented semantic segmentation module as illustrated in the blue block of Figure \ref{fig:architecture}. In particular, given a query point $q$, we have its neighbouring points $\{p_1 \dots p_k \dots p_K\}$ and point features $\{\boldsymbol{F}_1 \dots \boldsymbol{F}_k \dots \boldsymbol{F}_K\}$ at hand. Our module only takes into account the information of these neighbouring points to infer the semantic class, while ignoring the absolute position information of query point $q$. Formally, our module learns the semantic class for point $q$ as follows:

\begin{equation}
    s_q = MLPs\bigg(\mathcal{A} \Big([p_1 \oplus \boldsymbol{F}_1]\dots \dots [p_K \oplus \boldsymbol{F}_K] \Big) \bigg)
    \label{eq:semantic}
\end{equation}
where $\mathcal{A}$ is also Attsets, $p_1 \cdots p_K$ are the neighbouring point positions concatenated for training stability.

Fundamentally, our above formulation aims to learn a semantic class for the surface patch formed by the $K$ neighbouring points, instead of for the query point $q$ directly. As illustrated in Figure \ref{fig:ambiguity_semantic}, given the same surface patch formed by neighbouring points $\{p_1,p_2,p_3\}$, for all query points $\{q_1,q_2,\dots\}$ near such surface patch, our surface-oriented segmentation module is driven to learn a consistent semantic class, thus eliminating the sensitivity caused by the absolute position of query point $q$. 

\subsection{End-to-End Implementation}

\phantom{xxx}\textbf{Training:}
Our entire framework is trained end-to-end from scratch without any pretraining. The unsigned distance is optimized using $\ell_1$ loss and the semantic segmentation using cross-entropy loss $\ell_{ce}$. To avoid manually tuning the weights between two losses for experiments in Sec \ref{sec:sem_rec}, we apply the uncertainty loss \cite{Kendall2018} with default settings. 
The number of nearest neighbours $K$ is set as 4 in all experiments. ADAM optimizer with default parameters is adopted and the learning rate is set as $10^{-3}$ in all epochs.

\textbf{Explicit Semantic Surfaces Extraction:} In testing, given sparse point clouds as the input, we use the same algorithm introduced in NDF \cite{Chibane2020a} to extract dense point clouds and the Marching Cubes to extract meshes together with semantics for evaluation. Other details are in appendix. 

\section{Experiments}

\subsection{Overview}
We evaluate our \nickname{} in two categories of experiments. First, we evaluate the accuracy of surface reconstruction on four point cloud datasets, including Synthetic Rooms~\cite{Peng2020a}, ScanNet~\cite{Dai2017}, 2D-3D-S~\cite{Armeni2017} and SceneNN~\cite{Hua2016}. Note that, only Synthetic Rooms consists of closed 3D surfaces, while the other three are real-world datasets with complex topology and noisy open surfaces. Second, we jointly evaluate both semantic segmentation and surface reconstruction of our \nickname{} on the three challenging real-world datasets, and extensively investigate how one task might benefit the other. For all datasets, we follow their original train/val/test splits. More details are in appendix.

\textbf{Training Data Generation:}
For all datasets, we follow the same pre-processing steps used in NDF~\cite{Chibane2020a} and ConvOcc~\cite{Peng2020a} to normalize each ground truth scene mesh into a unit cube. For each scene, we sample both on and off surface points as the query points in training. For each query point, we find its nearest face in the ground truth mesh, and then calculate the unsigned distance value. Naturally, we directly assign the semantic label of the nearest face to that query point. With the $xyz$ positions of all query points and their unsigned distances and semantics, we train our \nickname{} in an end-to-end fashion.

\textbf{Metrics:}
To evaluate the accuracy of reconstruction, we use the standard Chamfer-$L_1$ Distance (CD-$L_1$$\times 10^{-2}$, $\downarrow$), Chamfer-$L_2$ Distance (CD-$L_2$$\times 10^{-4}$, $\downarrow$) and F-score ($\uparrow$) with different thresholds (FS-$\delta$, FS-2$\delta$, FS-4$\delta$, $\delta$=0.005) as primary metrics \cite{Peng2020a}. To evaluate the performance of semantic segmentation, we report the standard metrics including the mean IoU (mIoU, $\uparrow$) and Overall Accuracy (OA, $\uparrow$) of all classes. Following NDF and ConvOcc, all scores are computed by comparing the point clouds sampled from predicted implicit surfaces and ground truth meshes. 

\begin{table*}[th]
    \centering
     \resizebox{1.0\linewidth}{!}{
\begin{tabular}{r||cccc|cccc|cccc}
     & \multicolumn{4}{c|}{SceneNN}  & \multicolumn{4}{c|}{ScanNet}   & \multicolumn{4}{c}{2D-3D-S}                     \\ \hline \hline
Metrics   & CD-$L_1$  & CD-$L_2$  & FS-$\delta$   & FS-2$\delta$  & CD-$L_1$  & CD-$L_2$  & FS-$\delta$   & FS-2$\delta$  & CD-$L_1$    & CD-$L_2$  & FS-$\delta$   & FS-2$\delta$   \\ \hline 

NDF  & 0.460  & 0.248     & 0.726   & 0.927  & 0.385 & 0.214  & 0.800     & 0.964    & 0.418    & 0.523  & 0.762    & 0.969     \\
\textbf{Ours} & \textbf{0.327} & \textbf{0.169} & \textbf{0.834} & \textbf{0.977} & \textbf{0.286} & \textbf{0.125} & \textbf{0.884} & \textbf{0.988} & \textbf{0.327} & \textbf{0.194} & \textbf{0.845} & \textbf{0.977} \\  \hline
\end{tabular}
}
\vspace{0.2cm}
    \caption{Quantitative results of our \nickname{} and NDF on three real-world datasets: SceneNN, ScanNet and 2D-3D-S.}
	\label{tbl:recon_real_quan}
\vspace{-1.2cm}
\end{table*}

\begin{table*}[th]
    \centering
     \resizebox{1.0\linewidth}{!}{
\begin{tabular}{r||cccc|cccc|cccc}
Trained on             & CD-$L_1$       & CD-$L_2$       & FS-$\delta$       & FS-2$\delta$        & CD-$L_1$       & CD-$L_2$       & FS-$\delta$       & FS-2$\delta$        & CD-$L_1$       & CD-$L_2$       & FS-$\delta$       & FS-2$\delta$        \\ \hline \hline
Synthetic:  & \multicolumn{4}{c|}{Tested on SceneNN}                                      & \multicolumn{4}{c|}{Tested on ScanNet}                                      & \multicolumn{4}{c}{Tested on 2D-3D-S}                                       \\ \hline
ConvOcc               & 0.816          & 1.733          & 0.421          & 0.786          & 0.845          & 1.902          & 0.397          & 0.778          & 0.960          & 2.433          & 0.323          & 0.884          \\
NDF                  & 0.455          & 0.286          & 0.649          & 0.962          & 0.452          & 0.281          & 0.648          & 0.960          & 0.468          & 0.286          & 0.609          & 0.969          \\
SA-Conv             & 0.744          & 1.223          & 0.393          & 0.836          & 0.776          & 1.662          & 0.346          & 0.833          & 0.874          & 1.983          & 0.303          & 0.811          \\
\textbf{Ours} & \textbf{0.332} & \textbf{0.176} & \textbf{0.827} & \textbf{0.975} & \textbf{0.303} & \textbf{0.139} & \textbf{0.864} & \textbf{0.986} & \textbf{0.327} & \textbf{0.160} & \textbf{0.838} & \textbf{0.981} \\ \hline \\
SceneNN:          & \multicolumn{4}{c|}{Tested on Synthetic Rooms}                  & \multicolumn{4}{c|}{Tested on ScanNet}                                      & \multicolumn{4}{c}{Tested on 2D-3D-S}                                       \\ \hline
NDF                  & 0.569          & 0.458          & 0.404          & 0.868          & 0.462          & 0.389          & 0.707          & 0.928          & 0.688          & 1.712          & 0.662          & 0.858          \\
\textbf{Ours} & \textbf{0.474} & \textbf{0.407} & \textbf{0.627} & \textbf{0.904} & \textbf{0.285} & \textbf{0.127} & \textbf{0.880} & \textbf{0.989} & \textbf{0.340} & \textbf{0.190} & \textbf{0.826} & \textbf{0.977} \\ \hline \\
ScanNet:          & \multicolumn{4}{c|}{Tested on Synthetic Rooms }                 & \multicolumn{4}{c|}{Tested on SceneNN}                                      & \multicolumn{4}{c}{Tested on 2D-3D-S}                                       \\ \hline
NDF                  & 0.568          & 0.431          & 0.401          & 0.881          & 0.425          & 0.273          & 0.730          & 0.948          & 0.442          & 0.284          & 0.698          & 0.948          \\
\textbf{Ours} & \textbf{0.481} & \textbf{0.489} & \textbf{0.607} & \textbf{0.915} & \textbf{0.324} & \textbf{0.166} & \textbf{0.837} & \textbf{0.978} & \textbf{0.329} & \textbf{0.164} & \textbf{0.834} & \textbf{0.980} \\ \hline \\
2D-3D-S:          & \multicolumn{4}{c|}{Tested on Synthetic Rooms}           & \multicolumn{4}{c|}{Tested on SceneNN}                                      & \multicolumn{4}{c}{Tested on ScanNet}                                       \\ \hline
NDF                  & 0.527          & 1.799          & 0.645          & 0.972          & 0.382          & 0.217          & 0.780          & 0.970          & 0.378          & 0.205          & 0.787          & 0.972          \\
\textbf{Ours} & \textbf{0.432} & \textbf{0.310} & \textbf{0.654} & \textbf{0.929} & \textbf{0.314} & \textbf{0.161} & \textbf{0.845} & \textbf{0.978} & \textbf{0.272} & \textbf{0.112} & \textbf{0.898} & \textbf{0.991} \\ \hline
\end{tabular}
}
\vspace{0.3cm}
    \caption{Quantitative results of our method and baselines in the generalization of surface reconstruction across four datasets.}
	\label{tbl:exp_cross_quan}
\vspace{-1.2cm}
\end{table*}

\begin{table}[t]
    \centering
     \resizebox{0.65\linewidth}{!}{
        \begin{tabular}{r||ccccc}
        Methods     & CD-$L_1$  & CD-$L_2$ & FS-$\delta$ & FS-2$\delta$ & FS-4$\delta$ \\ \hline \hline
        SPSR             & 2.083          & -  & -     & 0.762          & 0.812 \\
        Trimmed SPSR        & 0.690          & -  & -      & 0.892          & -     \\
        PointConv           & 1.650          & -  & -      & 0.790          & -     \\
        OccNet          & 2.030          & - & -       & 0.541          & -     \\
        SAL                    & 2.720          & -  & -      & 0.405          & 0.598 \\
        IGR                & 1.923          & -   & -     & 0.740          & 0.812 \\
        LIG                   & 1.953          & -   & -     & 0.625          & 0.710 \\
        ConvOcc       & 0.420          & 0.538  & {\ul 0.778}  & {\ul 0.964}    & 0.983 \\
        NDF                   & {\ul 0.408}    & {\ul 0.301}  & 0.713 & 0.952 & {\ul 0.998}\\
        SA-CONet                  & 0.496          & 0.686    & 0.747 & 0.936        & 0.986   \\
        \textbf{\nickname{}}                        & \textbf{0.348} & \textbf{0.179} & \textbf{0.803} & \textbf{0.978} & \textbf{0.999} \\  \hline
        \end{tabular}
        }
    \vspace{0.4cm}
    \caption{Quantitative comparison of our \nickname{} with existing methods on scene-level reconstruction of Synthetic Rooms. The best scores are in bold and the second best are underlined.}
	\label{tbl:recon_syn_quan}
	\vspace{-.8cm}
\end{table}

\subsection{Surface Reconstruction}
To thoroughly evaluate our \nickname{}, we conduct two groups of experiments: 1) reconstruction on each of the four benchmark datasets, 2) generalization across unseen datasets. In all experiments, we follow the same settings of \cite{Mescheder2019,Chibane2020a,Peng2020a}. In particular, we use 10k on-surface points and 100k off-surface points of each scene in training. In testing, we randomly sample 100k points from the reconstructed surfaces to compute scores. All other details are provided in appendix.

\begin{figure*}[t]
\centering
   \includegraphics[width=1.\linewidth]{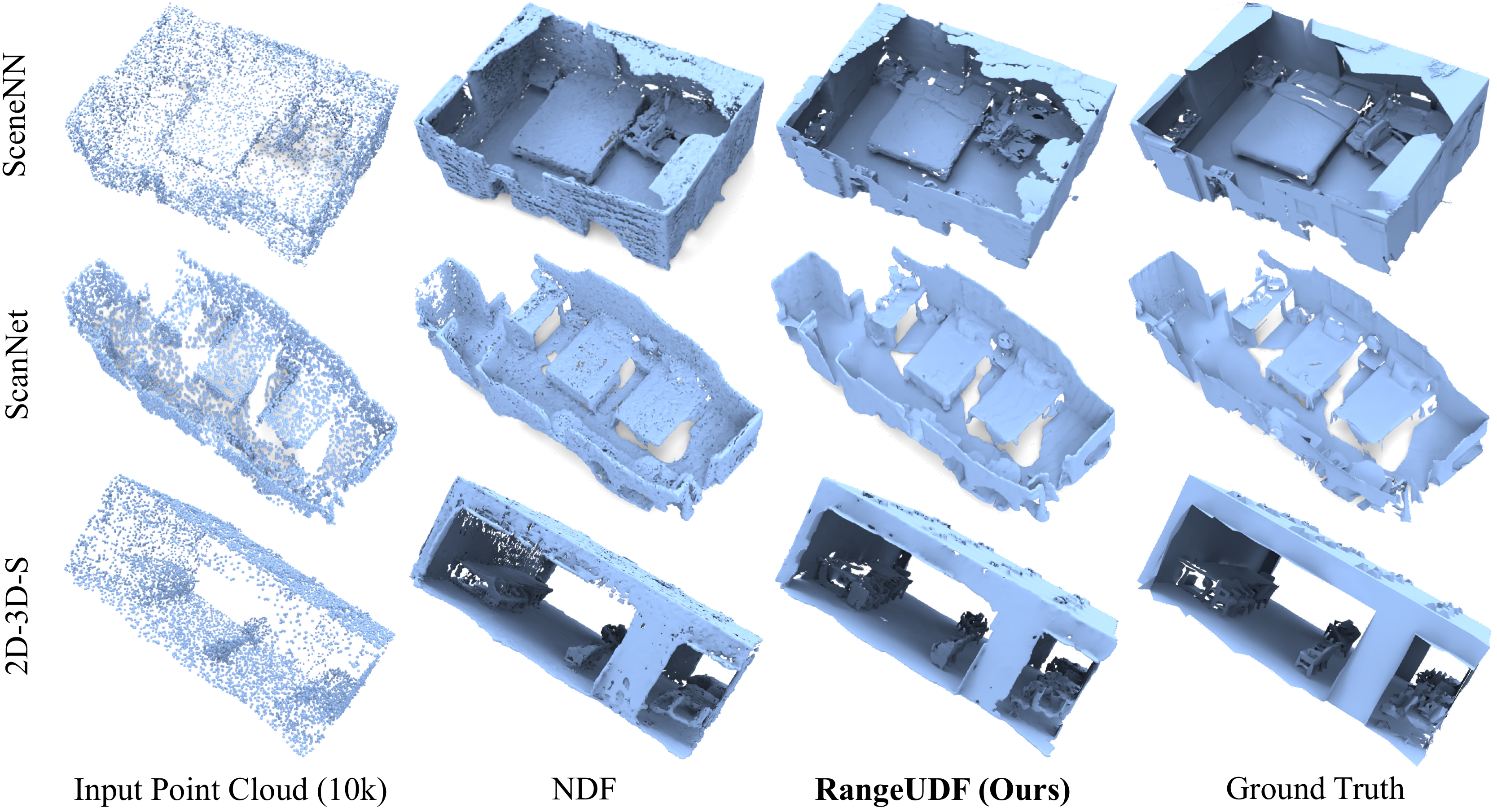}
\caption{Qualitative results of surface reconstruction from our \nickname{} and NDF on three real-world datasets: SceneNN, ScanNet and 2D-3D-S. For a fair comparison and visualization, we use the same level value to obtain the approximate meshes using Marching Cubes for both NDF and our RangeUDF.}
\label{fig:recon_self_qual}

\end{figure*}

\begin{figure*}[t]
\centering
   \includegraphics[width=1.\linewidth]{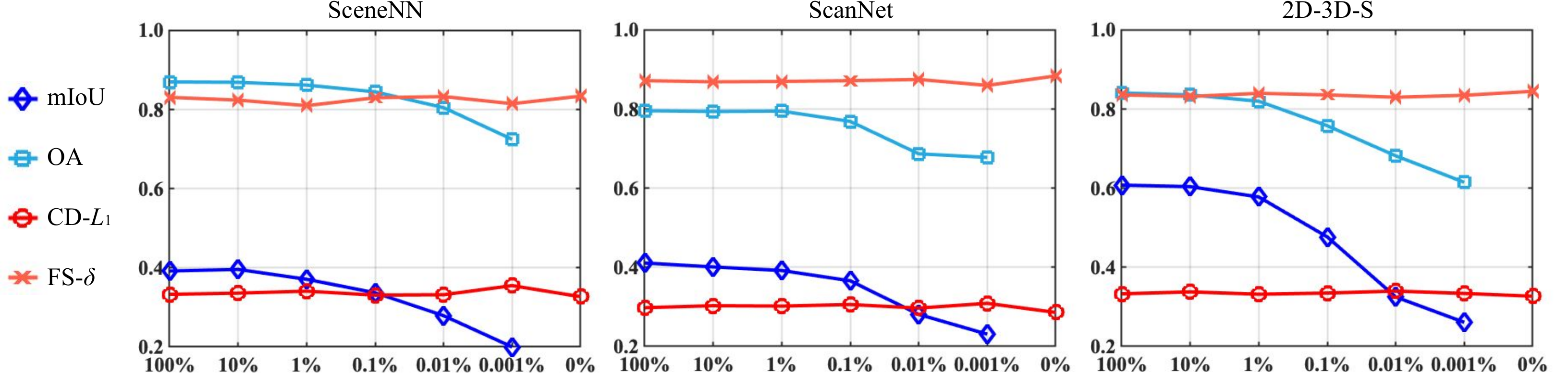}

\caption{Quantitative results of both surface reconstruction and semantic segmentation of our \nickname{} on the testing splits of three datasets giving different amounts of semantic training points. For comparison, the results of reconstruction only is also given at 0\%.}
\vspace{-.2cm}
\label{fig:exp_joint}

\end{figure*}

\textbf{Results on Four Benchmark Datasets:}
In this group of experiments, all methods are trained and tested within each of the four datasets. Table~\ref{tbl:recon_syn_quan} compares the quantitative results of our method and ten solid baselines on Synthetic Rooms~\cite{Peng2020a}. Since this synthetic dataset has perfect closed 3D surfaces, it is amenable to all types of implicit representations and classical methods. It can be seen that our \nickname{} clearly outperforms all existing methods in all metrics, pushing the accuracy to the next level.

Table~\ref{tbl:recon_real_quan} shows the quantitative results on the three challening real-world datasets: SceneNN~\cite{Hua2016}, ScanNet~\cite{Dai2017} and 2D-3D-S~\cite{Armeni2017}. Since these datasets only have open 3D surfaces for training, we can only compare with NDF \cite{Chibane2020a}, while other methods fail to be trained. It can be seen that our \nickname{} surpasses NDF by large margins on all three datasets over all metrics. This demonstrates the superiority of our simple range-aware unsigned distance function for recovering real-world complex 3D scenes with arbitrary topology. Figure~\ref{fig:recon_self_qual} shows the qualitative results, and we can see that our method successfully recover continuous and fine-grained scene geometries, while NDF generates surfaces with many holes and fails to  interpolate high-quality continuous geometries. 

\textbf{Generalization to Unseen Datasets:}
We further evaluate the generalization capability of our \nickname{} in surface reconstruction. In particular, we train \nickname{} on one specific dataset and then directly test it on the other three datasets. For comparison, we include ConvOcc \cite{Peng2020a}, NDF \cite{Chibane2020a}, SA-CONet \cite{Tang2021a} to conduct the generalization experiments from Synthetic Rooms to the other three datasets. For generalization from real-world datasets, we can only compare with NDF.

As shown in Table \ref{tbl:exp_cross_quan}, our \nickname{} significantly outperforms all methods in all settings of generalization experiments. Remarkably, the reconstruction performance of our method is extremely consistent and stable across multiple unseen datasets in the challenging generalization experiments. Note that, the state of the art implicit methods including ConvOcc, NDF and SA-CONet, all adopt trilinear interpolation to obtain the feature vectors for query points, while our method uses the proposed range-aware neural interpolation module. This clearly shows the superiority of our \nickname{}. 

\subsection{Semantic Segmentation and Reconstruction}\label{sec:sem_rec}
In addition to recovering accurate 3D surfaces from point clouds, our \nickname{} can also infer semantic classes for continuous surfaces, while the existing implicit representation based methods cannot. Although there are a plethora of semantic segmentation approaches \cite{Qi2016,Li2018f,Thomas2019} specially designed for discrete 3D point clouds, their experimental settings are vastly different from our \nickname{}. Therefore, it is hard and unfair to directly compare the performance on the online benchmark. In fact, our simple semantic branch does not target at achieving the best performance for a specific set of discrete points. Instead, we aim to demonstrate that the semantics of continuous surfaces can be effectively learned for our implicit representation.

In this section, we turn to evaluate how the semantics of our implicit representation can be effectively learned, and how the surface reconstruction and semantic segmentation affect each other in the joint framework. In particular, we conduct the following two groups of experiments on the three real-world datasets: SceneNN~\cite{Hua2016}, ScanNet~\cite{Dai2017} and 2D-3D-S~\cite{Armeni2017}.

\vspace{0.2cm}
\textbf{Does semantic branch degrade surface reconstruction?}
\vspace{0.2cm}

\begin{figure}[t]
\centering
   \includegraphics[width=1.\linewidth]{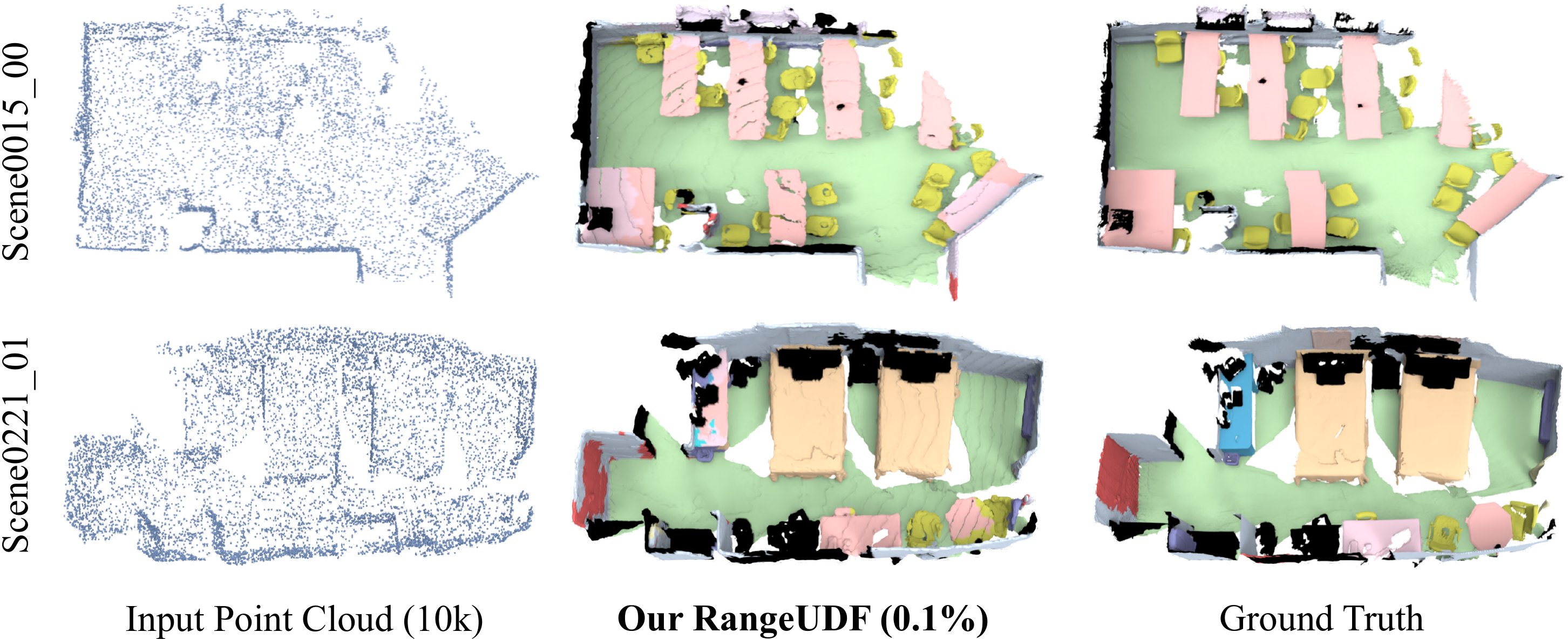}
\caption{Qualitative results of our method for joint 3D surface reconstruction and semantic segmentation on the validation split of ScanNet dataset.}
\label{fig:exp_joint_qualitative}
\vspace{-0.1cm}
\end{figure}

In this group of experiments, we simultaneously train our range-aware unsigned distance function and the surface-oriented semantic segmentation module with different amounts of semantic supervision signals. In particular, for each scene in the datasets, we sample 10k on-surface points and 100k off-surface points to train both branches in a fully-supervised fashion. For comparison, we train 5 additional groups of models, giving randomly sampled semantic annotations during training, ranging from 10\%, 1\%, 0.1\%, 0.01\%, to 0.001\%. Figure \ref{fig:exp_joint} shows the quantitative results of both surface reconstruction and semantic segmentation in the 6 different settings on three datasets, and Figure \ref{fig:exp_joint_qualitative} shows the qualitative results training with only 0.1\% of semantic labels. It can be seen that:

\begin{itemize}
    \item The accuracy of surface reconstruction is consistently superior even though the network is jointly trained with different amounts of semantic annotations. The CD-$L_1$ scores on all datasets only fluctuates within a margin of 0.024, and the FS-$\delta$ scores within a tiny range of 0.029. This shows that the high quality of surface reconstruction is barely influenced by semantic segmentation.
    \item Given as few as 1\% of full semantic annotations for training, the performance of our semantic segmentation only decreases by less than 3\% in mIOU scores compared with the model trained with 100\% annotations. This shows that our surface-oriented semantic segmentation module is robust to sparse semantic annotations.
\end{itemize}

\textbf{Does surface reconstruction benefit semantic branch?}
\vspace{0.2cm}

In this group of experiments, we aim to investigate whether our high quality surface reconstruction module can benefit the semantic branch. In particular, we simply remove the unsigned distance branch and train our network in a semantic-only mode (w/o Recon.) on three datasets. We then compare the semantic results with the models jointly trained with surface reconstruction in Figure \ref{fig:exp_joint}. 

Table \ref{tbl:exp_multitask_quan} compares the mIoU scores in different settings on three datasets. It can be seen that the semantic segmentation results can be consistently higher when the surface reconstruction branch is jointly optimized (w/ Recon mode), especially when the semantic annotations are scarce (e.g., $\leq 1\%$) during training. We hypothesize that the surface reconstruction module exhibits strong geometric priors such as continuity in spatial regions, which aids the network to propagate sparse semantic information to a wider context. 

\subsection{Ablation Study}
We conduct ablation studies to evaluate our framework. All ablated networks are evaluated on ScanNet~\cite{Dai2017} with 10k on-surface points and 100k off-surface points using full semantic labels. Since ScanNet scenes are spatially large, the voxel-based backbones of existing works such as NDF and ConvOcc are unable to be applied without time-consuming sliding windows. Therefore, we opt out ablation studies on voxel-based backbones.    
Because the range-aware neural interpolation and surface-oriented semantic segmentation modules are the major components of our \nickname{}, we conduct the following groups of ablation experiments. 
\begin{itemize}
    \item We only remove the explicit range-aware term $(q-p_k)$ in Eq.~\ref{eq:range} to evaluate the effectiveness of range information.
    \item We only add the position of the query point $q$ in our surface-oriented semantic branch in Eq.~\ref{eq:semantic}. 
    \item We choose different values for the hyperparameter $K$ in the KNN query step to verify its impact. 
    \item We only remove the uncertainty loss introduced to automatically balance the two branches. 
\end{itemize}

\begin{table}[t]
\centering
\resizebox{0.60\linewidth}{!}{
\begin{tabular}{r||cccccc}
Metric              & \multicolumn{6}{c}{mIoU}     \\ \hline \hline 
                & \multicolumn{2}{c|}{ScanNet} & \multicolumn{2}{c|}{2D-3D-S}  & \multicolumn{2}{c}{SceneNN} \\
Recon.            & w/o           & \multicolumn{1}{c|}{w/}            & w/o  & \multicolumn{1}{c|}{w/}            & w/o      & w/               \\ \hline
10\%                & \textbf{0.404} & \multicolumn{1}{c|}{0.401}          & 0.602 & \multicolumn{1}{c|}{\textbf{0.604}} & 0.393     & \textbf{0.396}    \\
1\%                 & 0.384          & \multicolumn{1}{c|}{\textbf{0.392}} & 0.567 & \multicolumn{1}{c|}{\textbf{0.568}} & 0.365     & \textbf{0.371}    \\
1\textperthousand    & 0.351          & \multicolumn{1}{c|}{\textbf{0.366}} & 0.473 & \multicolumn{1}{c|}{\textbf{0.477}} & 0.328     & \textbf{0.337}    \\
0.1\textperthousand  & 0.261          & \multicolumn{1}{c|}{\textbf{0.281}} & 0.304 & \multicolumn{1}{c|}{\textbf{0.325}} & 0.245     & \textbf{0.279}    \\
0.01\textperthousand & 0.205          & \multicolumn{1}{c|}{\textbf{0.231}} & 0.241 & \multicolumn{1}{c|}{\textbf{0.261}} & \textbf{0.184}        & 0.182 \\ \hline               
\end{tabular}
}
\vspace{0.4cm}
\caption{Quantitative results of semantic segmentation of our \nickname{} in different settings. Here, w/o and w/ denote that the framework is trained without and with reconstruction branch, respectively.}
\label{tbl:exp_multitask_quan}
\vspace{-.4cm}
\end{table}

\begin{table}[t]
    \centering
    \scriptsize
     \resizebox{0.6\linewidth}{!}{
\begin{tabular}{l|ccc}
Settings                                & CD-$L_1$       & FS-$\delta$    & mIoU     \\ \hline \hline
w/o $(q-p_k)$ in Eq.~\ref{eq:range}     & 0.324          & 0.856          & 0.407          \\
w/  $q$ in Eq.~\ref{eq:semantic}        & {\ul 0.300}    & {\ul 0.872}    & 0.392          \\
K=1                   & 0.313          & 0.850          & 0.396          \\
K=8                   & {\ul 0.300}    & {\ul 0.872}    & 0.400          \\
K=16                  & 0.305          & 0.866          & {\ul 0.409}    \\
w/o uncertainty loss      & 0.301          & 0.868          & 0.399          \\
\textbf{RangeUDF (Full)}                & \textbf{0.298} & \textbf{0.876} & \textbf{0.411} \\ \hline
\end{tabular}
}
\vspace{0.4cm}
    \caption{Quantitative results of ablated networks in semantic 3D surface reconstruction. Note that the results of our full framework is different from Table \ref{tbl:recon_real_quan} where the network is only trained for reconstruction.}
	\label{tbl:ablation}
\vspace{-.4cm}
\end{table}

From Table~\ref{tbl:ablation}, we can see that: 1) once the range-aware term $(q-p_k)$ is removed, the reconstruction performance decreases sharply and the CD-$L_1$ score is the worst, showing that adding this term, albeit technically simple, is crucial in our method; 2) once we add the position information of query point $q$ into semantic branch, the segmentation performance significantly drops and the mIoU score becomes the lowest, demonstrating that it is more effective to adopt our surface-orientated module; 3) given different choices of K with or without the uncertainty loss, the performance fluctuates within a reasonable range, showing the robustness of our framework overall.

\section{Limitations and Future Work}
Our \nickname{} is simple yet powerful to jointly reconstruct accurate 3D scene surfaces and estimate semantics from sparse point clouds. However, one limitation is the lack of object instance segmentation for our implicit representations. 
In addition, it is desirable to design a meshing strategy to extract accurate surfaces from the predicted unsigned distances instead of using Marching Cubes to find the approximate surfaces. It is also interesting to explore unsupervised learning techniques to automatically discover the surface semantics. 
We leave these problems for our future exploration.

\section{Conclusion}
In this paper, we propose \nickname{}, a simple and effective framework to simultaneously learn the structure and semantics of continuous surfaces from 3D point clouds. Our key components include a range-aware unsigned distance function which can estimate precise 3D structure without any surface ambiguity, and a surface-oriented semantic segmentation branch which can effectively learn semantics for implicit representations. Our \nickname{} demonstrates an unprecedented level of fidelity in 3D surface reconstruction, and has high potential to open up new opportunities in this area.

\clearpage
%
%
\bibliographystyle{splncs04}
\bibliography{references}

\appendix

\clearpage
\section{Appendix}
\subsection{Network Architecture}
\subsubsection{Feature Extractor}

\begin{figure}[h]
\centering
   \includegraphics[width=0.70\linewidth]{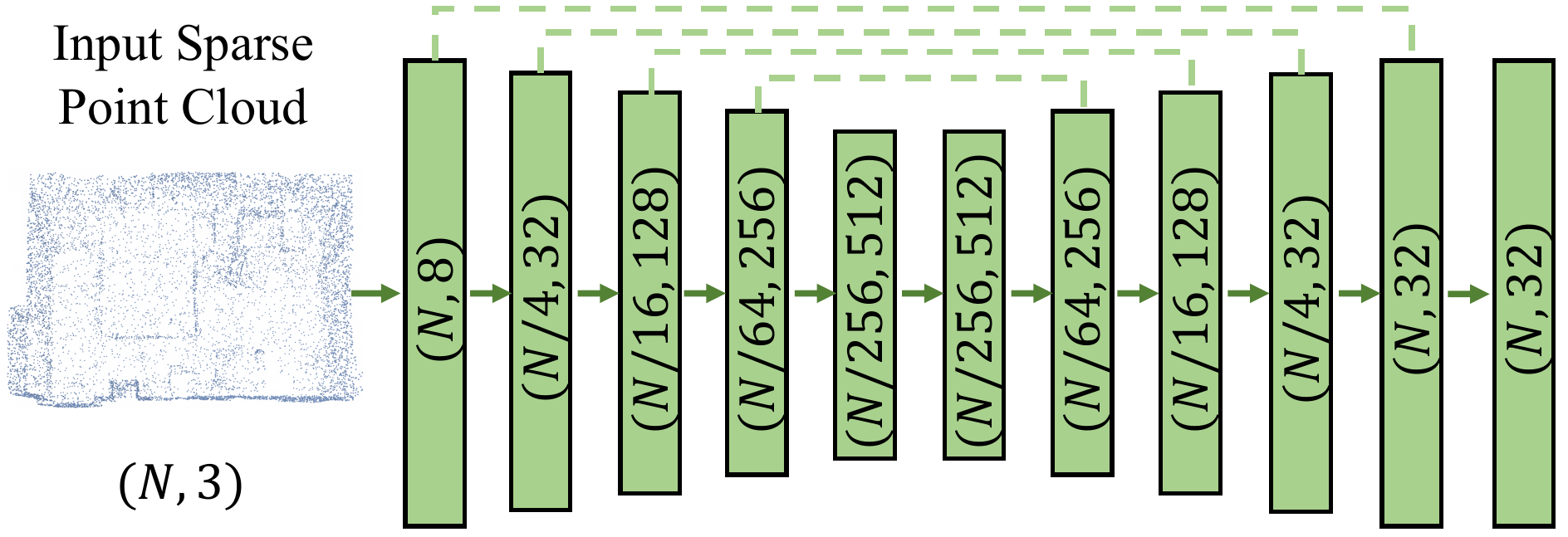}
\caption{The detailed architecture of feature extractor. We only modify the last layer of the decoder in RandLA-Net~\cite{Hu2020} to output a 32-D feature vector for each surface point.}
\label{fig:feat}
\vspace{-0.4cm}
\end{figure}

This module aims to extract per-point features from an input point cloud. As mentioned in Section \textcolor{red}{3.1}, we simply adopt the existing large-scale-point-cloud friendly RandLA-Net~\cite{Hu2020}. As shown in Figure~\ref{fig:feat}, given a raw point cloud with $N$ on-surface points $\{p_1 \dots p_n \dots p_N\}$ of a scene, a 4-level encoder-decoder with skip connections is applied to learn a 32-dimensional feature vector $\{\boldsymbol{F}_1 \dots \boldsymbol{F}_n \dots \boldsymbol{F}_N\}$ for each of $N$ points.

\begin{figure}[h]
\centering
   \includegraphics[width=0.9\linewidth]{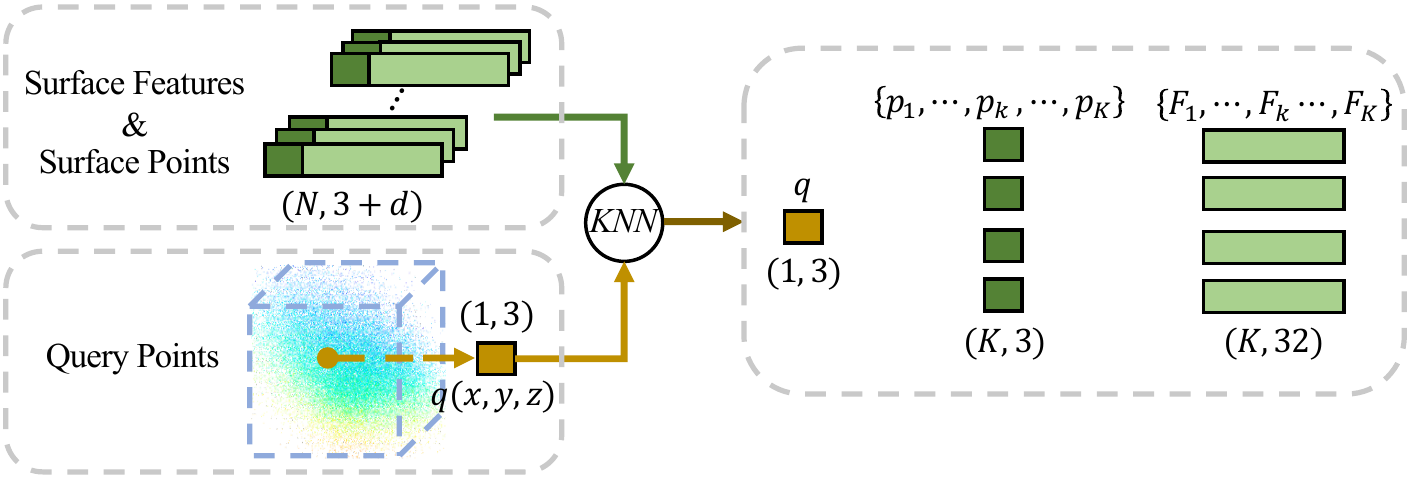}
\caption{The details of neighbourhood query module.}
\label{fig:knn}

\end{figure}

\subsubsection{Neighbourhood Query}

For the neighbourhood query module, we use kNN to collect K neighbouring points for every query point according to point Euclidean distances. As shown in Figure~\ref{fig:knn}, given a query point $q$, we first search the nearest $K$ points in $N$ surface points. Such $K$ neighbouring surface points $\{p_1 \dots p_k \dots p_K\}$  of $q$ and corresponding point features $\{\boldsymbol{F}_1 \dots \boldsymbol{F}_k \dots \boldsymbol{F}_K\}$ are retrieved.

\subsubsection{Range-aware Unsigned Distance Function}

\begin{figure}[h]
\centering
   \includegraphics[width=0.7\linewidth]{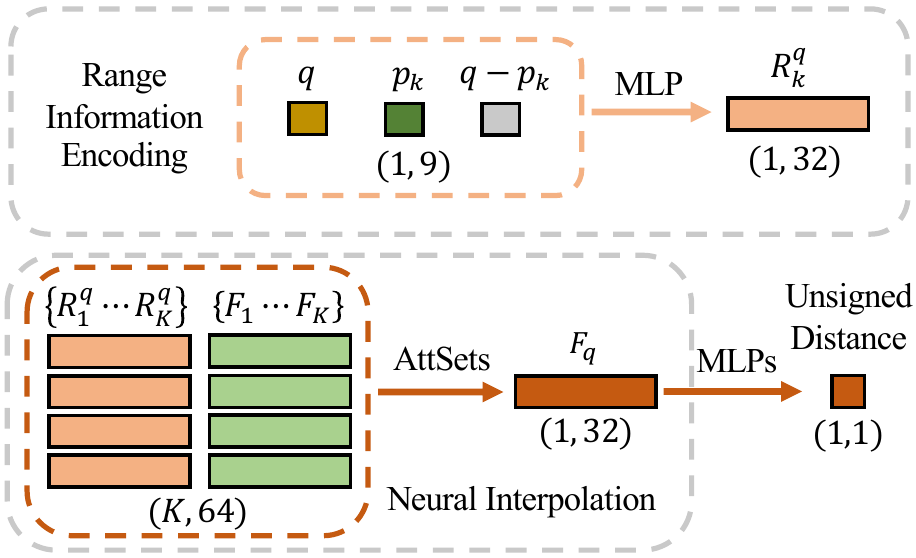}

\caption{The details of range-aware unsigned distance function.}

\label{fig:udf}
\end{figure}

Our range-aware unsigned distance function consists of: 1) range information encoding; 2) neural interpolation; and 3) unsigned distance regression.

1) Given a surface point $p_k$ as an example, we encode the range information for this neighbouring point as follows: 
\begin{equation}
    \boldsymbol{R}_k^q = MLP\Big(q \oplus p_k \oplus (q - p_k) \Big)
    \label{eq:range}
\end{equation}
where $q$ and $p_k$ are the $xyz$ positions of points, $\oplus$ is the concatenation operation. As shown in the top block in Figure~\ref{fig:udf}, the input of $MLP$ is a concatenated 9-dimensional position vector and the output is a 32-dimensional range vector $\boldsymbol{R}_k^q$. 

2) To interpolate a single feature vector $\boldsymbol{F}_u^q$ for the query point $q$, we concatenate the range vectors with point features followed by an attention pooling. Our neural interpolation is defined as follows:
\begin{equation}
    \boldsymbol{F}^q_u = \mathcal{A}\Big( [\boldsymbol{R}_1^q \oplus \boldsymbol{F}_1] \dots [\boldsymbol{R}_k^q \oplus \boldsymbol{F}_k] \dots [\boldsymbol{R}_K^q \oplus \boldsymbol{F}_K] \Big)
\end{equation}
where $\mathcal{A}$ is the simple AttSets \cite{Yang2020} in our experiments. As shown in the bottom block in Figure~\ref{fig:udf}, the input of AttSets is $K$ concatenated 64-dimensional vectors and the output is a 32-dimensional feature vector $\boldsymbol{F}^q_u$.
\begin{figure}[h]
\centering
   \includegraphics[width=0.75\linewidth]{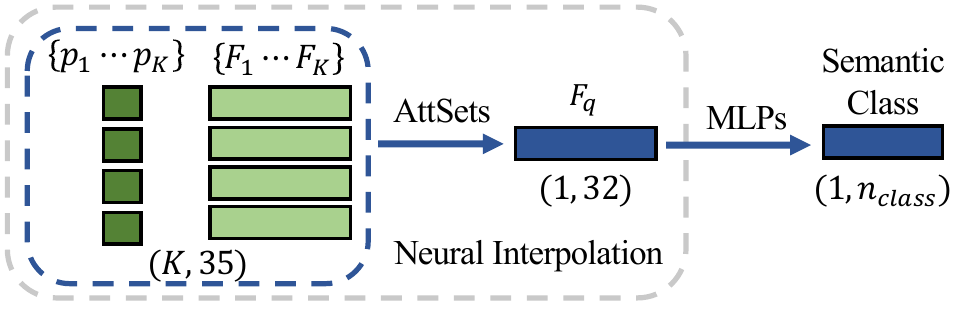}
   
\caption{The details of surface-oriented semantic segmentation.}

\label{fig:sem}
\end{figure}

3) Lastly, we directly feed the feature vector $\boldsymbol{F}^q_u$ of query point $q$ into 4 MLPs. The output dimensions of these MLPs are $(512\rightarrow32\rightarrow32\rightarrow1)$. For the first three MLPs, a LeakyReLU (slope=0.2) is integrated into each layer. The last MLP is followed by a ReLU function, enabling the distance value to be equal/greater than $0$.

\subsubsection{Surface-oriented Semantic Segmentation}

Our surface-oriented semantic segmentation module contains two modules: 1) surface-oriented interpolation and 2) semantic classification.

1) Given a query point $q$, we have its neighbouring points $\{p_1 \dots p_K\}$ and point features $\{\boldsymbol{F}_1 \dots \boldsymbol{F}_K\}$ at hand. Our module only takes into account the information of these neighbouring points. Formally, we learn the semantic feature for point $q$ as follows:
\begin{equation}
    \boldsymbol{F}^q_s = \mathcal{A} \Big([p_1 \oplus \boldsymbol{F}_1]\dots \dots [p_K \oplus \boldsymbol{F}_K] \Big)
    \label{eq:semantic}
\end{equation}
where $\mathcal{A}$ is also the attention function Attsets \cite{Yang2020} to aggregate the $K$ feature vectors. Specifically, the input of AttSets is $K$ concatenated 35-dimensional vectors and the output is a 32-dimensional semantic feature vector $\boldsymbol{F}^q_s$.

2) Then, we predict the semantic class for the query point $q$ from its semantic feature vector $\boldsymbol{F}^q_s$ by 3 MLPs. The output dimensions of these MLPs are $(64\rightarrow32\rightarrow n_{class})$. For the first two MLPs, a LeakyReLU (slope=0.2) is integrated into each layer.

\subsection{Data Preparation}
\subsubsection{Datasets} 
In this paper, we consider four point cloud datasets: Synthetic Rooms, ScanNet, 2D-3D-S and SceneNN.

\textbf{Synthetic Rooms}~\cite{Peng2020a} is a synthetic indoor dataset, consisting of 5000 scenes (3,750 for training, 250 for validation and 1,000 for testing). Each scene has several objects (chair, sofa, lamp, cabinet, table) from ShapeNet~\cite{Chang2015} . We follow the same split setting in ~\cite{Peng2020a} and use the whole test set to conduct quantitative evaluation. 

\textbf{SceneNN}~\cite{Hua2016} is an RGB-D dataset with 76 indoor scenes for the task of 3D semantic segmentation. There are 56 scenes for training and 20 scenes for testing~\cite{Hua2018} with 11 semantic classes. We adopt the same split setting in our experiments. 

\textbf{ScanNet}~\cite{Dai2017} contains 1,513 real-world rooms collected by an RGB-D camera. There are 20 semantic classes in the evaluation of 3D semantic segmentation. In particular, there are 1,201 scans for training and 312 for validation. Since ScanNet does not provide an online benchmark for surface reconstruction, we use the validation as our testing set and directly sample surface points from the provided raw (without alignment) non-watertight meshes.

\textbf{2D-3D-S}~\cite{Armeni2017} covers 6 large-scale indoor areas with 271 rooms  (Area-1: 44, Area-2: 40, Area-3: 23, Area4: 49, Area-5: 67, Area-6: 48) captured by Matterport sensors. There are 13 annotated semantic classes for this dataset. A non-watertight mesh is provided for each room. Note that, Area-5 is split into Area-5a and Area-5b, in which 47 rooms are unevenly broken into two parts. To avoid the imbalanced data introduced by Area-5, we choose Area-1$\sim$ Area-4 as our training set and Area-6 as the testing set. 

\subsubsection{Data Generation}
For all datasets, we follow the same pre-processing steps used in ConvOcc~\cite{Peng2020a} to normalize each ground truth scene mesh into a unit cube. For each scene, we randomly sample 10k surface points from the normalized mesh. For each surface point, we take the semantic class of the face that the surface point belongs to as its semantic class. Moreover, for all surface points, their unsigned distance values are all $0$.

We also sample 100k off-surface points in a unit cube for each scene using the same strategy in NDF~\cite{Chibane2020a}. For each off-surface point, we find its nearest face on the ground truth mesh and then calculate the corresponding unsigned distance value. Naturally, we directly assign the semantic label of the nearest face to that query point. It is noted that all surface and off-surface points are preserved and fixed for both training and testing after the sampling.

\subsection{Experiment Details}

We implement our \nickname{} with PyTorch. All experiments in the main paper are conducted on the same machine with an Intel(R) Xeon(R) E5-2698 v4 @ 2.20GHz CPU and an NVIDIA Tesla V100 GPU. Note that, for a fair comparison with ConvOcc~\cite{Peng2020a} and SA-ConvOnet~\cite{Tang2021a}, the evaluations are based on their provided pretrained models. In addition, as NDF does not conduct scene-level surface reconstruction in the original paper~\cite{Chibane2020a}, we carefully adapt it to our context based on the official implementation.

\subsubsection{Training}

During training, we use a batch size of 4 on all datasets. For each scene in the batch, we feed a point cloud with 10k points into the feature extractor, and feed randomly sampled 50k query points into the neighbourhood query module. 
We observe that our method can be quickly optimized. 
In particular, for the task of surface reconstruction on the ScanNet dataset, NDF~\cite{Chibane2020a} requires around 48.2 hours (85 epochs) to fully converge. In contrast, our \nickname{} only uses $\sim$10.4 hours (390 epochs). For an easy reproduction of our results in semantic surface reconstruction, we uniformly train our RangeUDF by 20 hours on each dataset which ensures the convergence for all datasets.
\begin{figure*}[t]
\centering
   \includegraphics[width=1.0\linewidth]{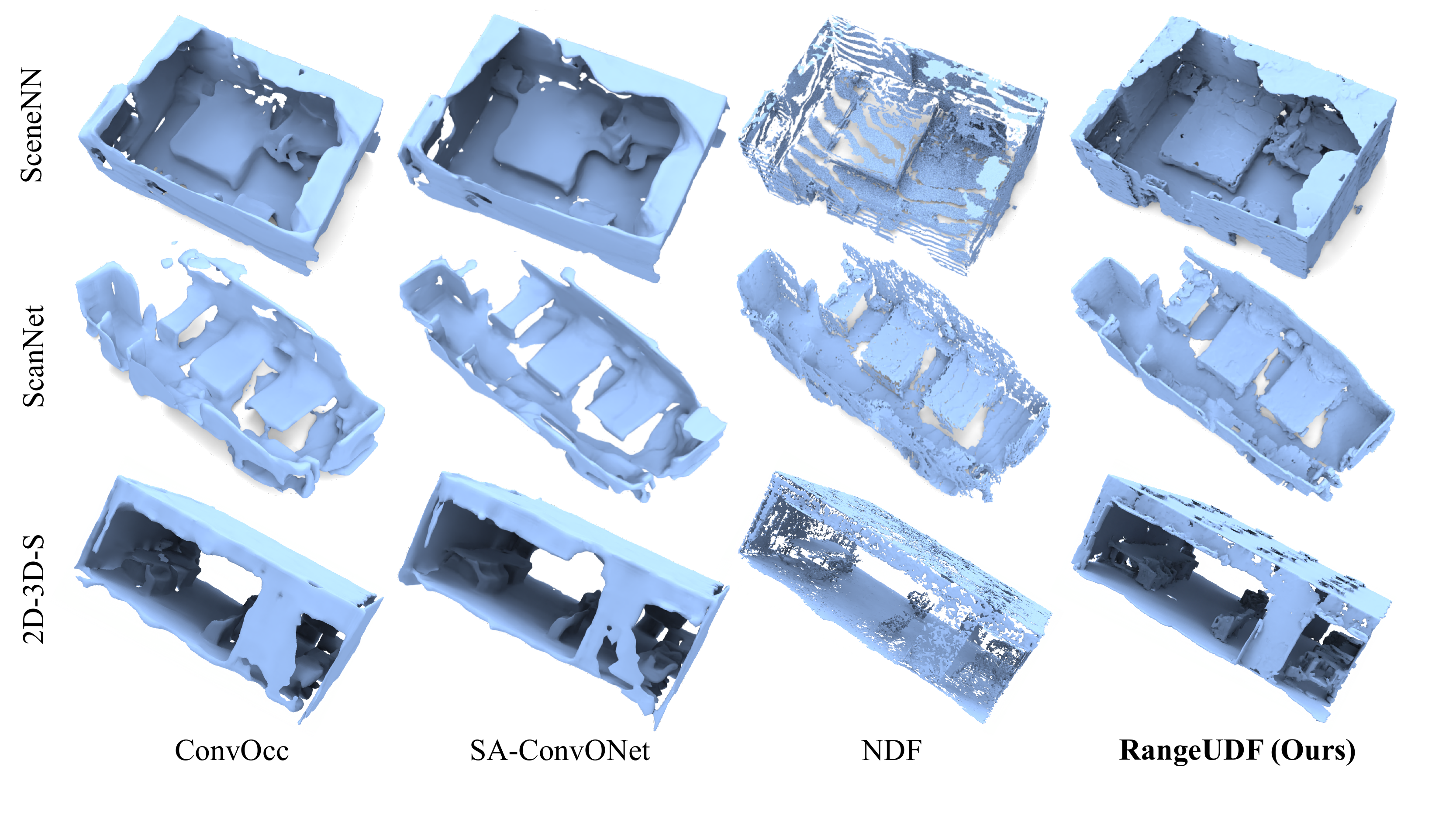}

\caption{Qualitative results of our method and baselines (ConvOcc~\cite{Peng2020a}, SA-ConvONet~\cite{Tang2021a} and NDF~\cite{Chibane2020a}) in the generalization of surface reconstruction from Synthetic Rooms~\cite{Peng2020a} to real-world SceneNN~\cite{Hua2016}, ScanNet~\cite{Dai2017} and 2D-3D-S~\cite{Armeni2017} datasets. For comparisons, all meshes  are obtained using Marching Cubes with the same settings.}
\label{fig:syn_real}

\end{figure*}
\subsubsection{Quantitative Evaluation}

To quantitatively compare the performance of our approach and baselines, we randomly sample 100k points (the same as ~\cite{Mescheder2019,Chibane2020a,Peng2020a}) from the reconstructed surfaces to compute both reconstruction metrics (CD-$L_1$, CD-$L_2$, F-score) and semantic segmentation metrics (mIoU, OA). Different from baselines such as OccNet, ConvOcc and SA-ConvONet which sample surface points from generated meshes, both NDF and our method sample surface points from extracted dense point clouds. 

For dense point cloud extraction, we use Algorithm 1 from NDF. Specifically, the distance threshold and the minimum number of generated points are set to 0.1 and 1,600k, respectively. Our \nickname{} consumes around 9.8s on average to generate a dense point cloud for a scene. However, NDF needs 60.2s for one scene. Additionally, only 0.8s is needed by our \nickname{} to infer the semantic classes for a generated dense point cloud (over 1,600k points).

\subsubsection{Qualitative Evaluation}

To comprehensively demonstrate the performance of our proposed approach, qualitative results are shown by generating meshes for each scene. For both NDF and our \nickname{}, we simply regress the unsigned distance value for each voxel in a volume at the resolution of $256^3$. To generate such volume, our method takes about 0.95s while NDF needs around 15.0s. The volume is then directly fed into the $marching\_cubes\_lewiner$ method from $skimage.measure$
with the setting of (level=$0.003$, spacing=$[1.0/255]*3$) to extract the mesh.

\subsection{Additional Results}

\subsubsection{Surface Reconstruction}

In Figure~\ref{fig:syn_real}, we also provide more qualitative results of generalization experiments from Synthetic Rooms~\cite{Peng2020a} to the other three datasets: SceneNN~\cite{Hua2016}, ScanNet~\cite{Dai2017} and 2D-3D-S~\cite{Armeni2017}. For comparison, we include the results from ConvOcc \cite{Peng2020a}, NDF~\cite{Chibane2020a}, SA-ConvONet \cite{Tang2021a}. To better demonstrate the generalization capability, we use the same scenes from qualitative results in Section \textcolor{red}{4.2}.

In particular, all methods are trained on the synthetic dataset and then directly test it on the other three real-world datasets. As shown in Figure~\ref{fig:syn_real}, our \nickname{} demonstrates significantly finer surface details, while ConvOcc / SA-ConvONet can only recover over-smooth surfaces and NDF fails to generalize to novel datasets. Remarkably, the qualitative results of our method is extremely consistent and stable across multiple unseen datasets. 
\begin{table*}[t]
\centering
\resizebox{\linewidth}{!}{
\begin{tabular}{r||cccccccc||cccccccc}
Datasets            & \multicolumn{8}{c||}{2S-3D-S}                                                                                                                                                                     & \multicolumn{8}{c}{SceneNN}                                                                                                                                                                      \\ \hline \hline
Tasks               & \multicolumn{4}{c|}{Semantic Segmentation}                                                                           & \multicolumn{4}{c||}{Surface Reconstruction}                                                    & \multicolumn{4}{c|}{Semantic Segmentation}                                                                           & \multicolumn{4}{c}{Surface Reconstruction}                                                     \\
Metrics             & \multicolumn{2}{c|}{OA (\%)}                       & \multicolumn{2}{c|}{mIoU (\%)}                     & \multicolumn{2}{c|}{CD-$L_1$}                   & \multicolumn{2}{c||}{FS-$\delta$}   & \multicolumn{2}{c|}{OA (\%)}                       & \multicolumn{2}{c|}{mIoU (\%)}                     & \multicolumn{2}{c|}{CD-$L_1$}                   & \multicolumn{2}{c}{FS-$\delta$}    \\
Points      & 10K           & \multicolumn{1}{c|}{50K}           & 10K           & \multicolumn{1}{c|}{50K}           & 10K            & \multicolumn{1}{c|}{50K}            & 10K            & 50K            & 10K           & \multicolumn{1}{c|}{50K}           & 10K           & \multicolumn{1}{c|}{50K}           & 10K            & \multicolumn{1}{c|}{50K}            & 10K            & 50K            \\ \hline
0.01\textperthousand & 61.5          & \multicolumn{1}{c|}{65.2}          & 26.1          & \multicolumn{1}{c|}{30.6}          & 0.334          & \multicolumn{1}{c|}{0.315}          & 0.835          & 0.857          & 72.5        & \multicolumn{1}{c|}{75.9}           & 18.2           & \multicolumn{1}{c|}{21.5}              & 0.355                & \multicolumn{1}{c|}{0.304}               & 0.815               & 0.853               \\
0.1\textperthousand  & 68.2          & \multicolumn{1}{c|}{76.6}          & 32.5          & \multicolumn{1}{c|}{46.4}          & 0.340          & \multicolumn{1}{c|}{\textbf{0.312}} & 0.830          & \textbf{0.869} & 80.5          & \multicolumn{1}{c|}{84.3}          & 27.9          & \multicolumn{1}{c|}{31.7}          & {\ul 0.332}    & \multicolumn{1}{c|}{0.310}          & \textbf{0.833} & 0.851          \\
1\textperthousand    & 75.8          & \multicolumn{1}{c|}{83.2}          & 47.7          & \multicolumn{1}{c|}{61.8}          & 0.335          & \multicolumn{1}{c|}{{\ul 0.314}}    & {\ul 0.836}    & {\ul 0.867}    & 84.5          & \multicolumn{1}{c|}{86.3}          & 33.7          & \multicolumn{1}{c|}{40.4}          & \textbf{0.331} & \multicolumn{1}{c|}{{\ul 0.299}}    & {\ul 0.830}    & {\ul 0.866}    \\
1\%                 & 82.0          & \multicolumn{1}{c|}{{\ul 86.3}}    & 56.8          & \multicolumn{1}{c|}{{\ul 66.5}}    & \textbf{0.332} & \multicolumn{1}{c|}{0.315}          & \textbf{0.840} & 0.864          & 86.2          & \multicolumn{1}{c|}{\textbf{89.1}} & 37.1          & \multicolumn{1}{c|}{43.2}          & 0.341          & \multicolumn{1}{c|}{0.303}          & 0.810          & 0.863          \\
10\%                & {\ul 83.6}    & \multicolumn{1}{c|}{{\ul 86.3}}    & {\ul 60.4}    & \multicolumn{1}{c|}{\textbf{67.7}} & 0.338          & \multicolumn{1}{c|}{0.315}          & 0.832          & 0.860          & {\ul 86.9}    & \multicolumn{1}{c|}{{\ul 87.9}}    & \textbf{39.6} & \multicolumn{1}{c|}{{\ul 43.4}}    & 0.336          & \multicolumn{1}{c|}{\textbf{0.294}} & 0.824          & \textbf{0.884} \\
100\%               & \textbf{84.1} & \multicolumn{1}{c|}{\textbf{86.7}} & \textbf{60.8} & \multicolumn{1}{c|}{{\ul 66.5}}    & {\ul 0.333}    & \multicolumn{1}{c|}{{\ul 0.314}}    & {\ul 0.836}    & 0.866          & \textbf{87.0} & \multicolumn{1}{c|}{{\ul 87.9}}    & {\ul 39.2}    & \multicolumn{1}{c|}{\textbf{43.8}} & 0.333          & \multicolumn{1}{c|}{0.303}          & 0.831          & 0.865         
\end{tabular}
}
\vspace{0.3cm}
\caption{Quantitative results of semantic surface reconstruction on 2D-3D-S~\cite{Armeni2017} and SceneNN~\cite{Hua2016}. For these two datasets, we evaluate the impact of two factors: 1) the percentage of points with semantic labels (varying from 0.01\% to 100\%); 2). the number of surface points (10K and 50K). The best results on different metrics are in bold and the second-best ones are underlined.}

\label{tbl:s3_scene}
 
\end{table*}
\subsubsection{Semantic Surface Reconstruction}

1) We evaluate the performance of semantic segmentation using the generated dense point clouds. We also calculate the results on ScanNet~\cite{Dai2017} using point clouds (all vertices) directly from ground truth meshes, and we get a 40.8\% mIoU. This is almost the same as our results in Table \textcolor{red}{6} 
(41.1\% mIoU). This shows that using the generated point clouds to evaluate the performance of semantic segmentation in our context is more valid and meaningful.

2) In Section \textcolor{red}{3.3}, we argue that optimizing the semantic segmentation branch with on-surface points only would result in imbalanced and ineffective optimization between reconstruction and semantic segmentation branches. To verify this, we modify the training strategy of our \nickname{}. In particular, we only use on-surface points to optimize the semantic branch on the ScanNet~\cite{Dai2017} dataset. Given the same generated dense point clouds, such a strategy achieves 39.1\% mIoU during inference. In contrast, 41.1\% mIoU is reported when both on/off-surface points are considered for semantic segmentation during training. 

3) We also report the detailed experimental results of semantic surface reconstruction on the SceneNN~\cite{Hua2016}, ScanNet~\cite{Dai2017} and 2D-3D-S~\cite{Armeni2017} datasets in Table~\ref{tbl:scannet_only} and Table~\ref{tbl:s3_scene}. As shown in these two tables, we additionally explore the impact of color and surface point density on the performance of surface reconstruction and semantic segmentation. 

\begin{table*}[t]
\centering
\resizebox{\linewidth}{!}{
\begin{tabular}{r||cccccccc||cccccccc}
Tasks               & \multicolumn{8}{c||}{Semantic Segmentation}                                                                                                                                                          & \multicolumn{8}{c}{Surface Reconstruction}                                                                                                                                                                   \\ \hline \hline
Color                 & \multicolumn{4}{c|}{w/o RGB}                                                                            & \multicolumn{4}{c||}{w/ RGB}                                                               & \multicolumn{4}{c|}{w/o RGB}                                                                                & \multicolumn{4}{c}{w/ RGB}                                                             \\
Metrics             & \multicolumn{2}{c|}{OA (\%)}                       & \multicolumn{2}{c|}{mIoU (\%)}                     & \multicolumn{2}{c|}{OA (\%)}                             & \multicolumn{2}{c||}{mIoU (\%)} & \multicolumn{2}{c|}{CD-$L_1$}                   & \multicolumn{2}{c|}{FS-$\delta$}                        & \multicolumn{2}{c|}{CD-$L_1$}                   & \multicolumn{2}{c}{FS-$\delta$}    \\
Points              & 10K           & \multicolumn{1}{c|}{50K}           & 10K           & \multicolumn{1}{c|}{50K}           & 10K                 & \multicolumn{1}{c|}{50K}           & 10K            & 50K           & 10K            & \multicolumn{1}{c|}{50K}            & 10K            & \multicolumn{1}{c|}{50K}            & 10K            & \multicolumn{1}{c|}{50K}            & 10K            & 50K            \\ \hline
0.01\textperthousand & 67.8          & \multicolumn{1}{c|}{71.5}          & 23.1          & \multicolumn{1}{c|}{28.4}          & 70.9                & \multicolumn{1}{c|}{73.4}          & 23.4           & 29.1          & 0.309          & \multicolumn{1}{c|}{{\ul 0.255}}    & 0.860          & \multicolumn{1}{c|}{0.925}          & 0.301          & \multicolumn{1}{c|}{0.258}          & 0.865          & 0.919          \\
0.1\textperthousand  & 68.7          & \multicolumn{1}{c|}{79.7}          & 28.1          & \multicolumn{1}{c|}{39.0}          & 74.1                & \multicolumn{1}{c|}{80.1}          & 30.9           & 41.7          & \textbf{0.297} & \multicolumn{1}{c|}{\textbf{0.253}} & {\ul 0.875} & \multicolumn{1}{c|}{{\ul 0.929}}    & 0.295          & \multicolumn{1}{c|}{0.262}          & 0.876          & 0.916          \\
1\textperthousand    & 76.8          & \multicolumn{1}{c|}{82.2}          & 36.6          & \multicolumn{1}{c|}{47.6}          & 79.6                & \multicolumn{1}{c|}{82.8}          & 39.5           & 48.7          & 0.306          & \multicolumn{1}{c|}{0.258}          & 0.872    & \multicolumn{1}{c|}{\textbf{0.930}} & {\ul 0.290}    & \multicolumn{1}{c|}{{\ul 0.251}}    & {\ul 0.881}    & {\ul 0.931}    \\
1\%                 & {\ul 79.5}    & \multicolumn{1}{c|}{83.1}          & 39.2          & \multicolumn{1}{c|}{50.0}          & 81.5                & \multicolumn{1}{c|}{82.5}          & 41.9           & 49.5          & 0.302          & \multicolumn{1}{c|}{0.260}          & 0.870          & \multicolumn{1}{c|}{0.917}          & \textbf{0.284} & \multicolumn{1}{c|}{0.268}          & \textbf{0.894} & 0.917          \\
10\%                & 79.4          & \multicolumn{1}{c|}{{\ul 83.2}}    & {\ul 40.1}    & \multicolumn{1}{c|}{{\ul 50.8}}    & {\ul \textbf{81.8}} & \multicolumn{1}{c|}{{\ul 83.8}}    & {\ul 42.7}     & {\ul 50.7}    & 0.303          & \multicolumn{1}{c|}{0.266}          & 0.869          & \multicolumn{1}{c|}{0.922}          & 0.296          & \multicolumn{1}{c|}{\textbf{0.248}} & 0.875          & \textbf{0.935} \\
100\%               & \textbf{79.6} & \multicolumn{1}{c|}{\textbf{83.5}} & \textbf{41.1} & \multicolumn{1}{c|}{\textbf{50.1}} & {\ul 81.6}          & \multicolumn{1}{c|}{\textbf{84.3}} & \textbf{44.0}  & \textbf{51.1} & {\ul 0.298}    & \multicolumn{1}{c|}{0.264}          & \textbf{0.876}    & \multicolumn{1}{c|}{0.912}          & 0.294          & \multicolumn{1}{c|}{0.261}          & 0.876          & 0.917         
\end{tabular}
}
\vspace{0.4cm}
\caption{Quantitative results of semantic surface reconstruction on the ScanNet~\cite{Dai2017}. We evaluate the impact of three factors: 1) the percentage of points with semantic labels (varying from 0.01\% to 100\%); 2) RGB information (w/o RGB and w/ RGB); 3). the number of surface points (10K and 50K). The best results on different metrics are in bold and the second-best ones are underlined.}

\label{tbl:scannet_only}
\end{table*}

More qualitative results can be found in the supplied video.

\end{document}